%% file: free_space_detection_for_autonomous_forklifts.tex
\documentclass[letterpaper, 10 pt, conference]{ieeeconf}
\IEEEoverridecommandlockouts
\usepackage{cite}
\usepackage{amsmath,amssymb,amsfonts}
\usepackage{algorithmic}
\usepackage{graphicx}
\usepackage{textcomp}
\usepackage{xcolor}
\usepackage[nolist]{acronym}

\usepackage{url}
\usepackage[hidelinks]{hyperref}
\usepackage[group-separator={,},output-decimal-marker={.},group-minimum-digits=4]{siunitx}
\usepackage{tabularx}
\usepackage{booktabs}
\usepackage{subcaption}
\usepackage{amsmath}
\usepackage{float}
\usepackage{pdfpages}
\usepackage{caption}
\newcommand{\norm}[1]{\left\lVert#1\right\rVert}
\newcommand{\kittimiou}{50.90}
\newcommand{\kittipercent}{\SI{\kittimiou}{\percent}}

\newcommand{\kittihz}{\num[round-mode=places, round-precision=0]{105.04}}
\newcommand{\kittisamples}{\num{23201}}
\newcommand{\fsdmiou}{63.30} 
\newcommand{\fsdpercent}{\SI{\fsdmiou}{\percent}} 
\newcommand{\fsdhz}{\num[round-mode=places, round-precision=0]{53.65}}

\newcommand{\fsdsamples}{\num{4464}}
\newcommand{\filtersigma}{1}

\def\BibTeX{{\rm B\kern-.05em{\sc i\kern-.025em b}\kern-.08em
    T\kern-.1667em\lower.7ex\hbox{E}\kern-.125emX}}
\begin{document}

\def\etal{{et al. }}
\def\example{{e.g. }}
\def\thatis{{i.e. }}

\makeatletter
\patchcmd{\maketitle}{\@fnsymbol}{\@alph}{}{}  
\makeatother

\title{%
    Height Change Feature Based Free Space Detection} 


\author{
    Steven Schreck\thanks{\textsuperscript{*} The authors contributed equally.}\textsuperscript{*}\thanks{\textsuperscript{1}The authors are with the Cooperative Automated Traffic Systems Department, University of Applied Sciences Aschaffenburg (Germany).
    E-mail: \{steven.schreck, hannes.reichert, manuel.hetzel, konrad.doll\}@th-ab.de.}\textsuperscript{1}
    \and
    Hannes Reichert\textsuperscript{*}\textsuperscript{1}
    \and
    Manuel Hetzel\textsuperscript{1}
    \and
    Konrad Doll\textsuperscript{1}
    \and
    Bernhard Sick\textsuperscript{2}\thanks{\textsuperscript{2}The author is with the AI for Computationally Intelligent Systems Department, Kassel University (Germany). E-mail: \{bsick\}@uni-kassel.de.}
    \thanks{This work results from the project KAnIS (Kooperative Autonome Intralogistik-Systeme) by ``FuE Programm Informations- und Kommunikationstechnik Bayern'', grant number DIK-1910-0016// DIK0103/01 and the industrial partner: Linde Material Handling GmbH. Additionally, the work is supported by the Federal Ministry for Economic Affairs and Climate Action (BMWK), grant numbers 19A20001L and 19A20001O.
    }
}

\maketitle
\begin{acronym}
    \acro{agv}[AGV]{Automated Guided Vehicle}
    \acro{lidar}[LiDAR]{Light Detection And Ranging}
    \acro{cnn}[CNN]{Convolutional Neural Network}
    \acro{roi}[ROI]{Region of Interest}
    \acro{ml}[ML]{Machine Learning}
    \acro{wod}[WOD]{Waymo Open Dataset}
    \acro{bev}[BEV]{Bird Eye View}
    \acro{mlp}[MLP]{Multilayer Perceptron}
    \acro{iou}[IoU]{Intersection over Union}
    \acro{map}[mAP]{mean Average Precision}
    \acro{hsv}[HSV]{Hue Saturation Value}
    \acro{miou}[mIoU]{mean Intersection over Union}
    \acro{fov}[FOV]{Field of View}
    \acro{rgpf}[R-GPF]{Region-wise Ground Plane Fitting}
\end{acronym}


\begin{abstract}
In the context of autonomous forklifts, ensuring non-collision during travel, pick, and place operations is crucial.
To accomplish this, the forklift must be able to detect and locate areas of free space and potential obstacles in its environment.
However, this is particularly challenging in highly dynamic environments, such as factory sites and production halls, due to numerous industrial trucks and workers moving throughout the area.
In this paper, we present a novel method for free space detection, which consists of the following steps. 
We introduce a novel technique for surface normal estimation relying on spherical projected \acs{lidar} data. 
Subsequently, we employ the estimated surface normals to detect free space.
The presented method is a heuristic approach that does not require labeling and can ensure real-time application due to high processing speed.
The effectiveness of the proposed method is demonstrated through its application to a real-world dataset obtained on a factory site both indoors and outdoors, and its evaluation on the \textit{Semantic KITTI dataset}~\cite{2019.semantic_kitti}.
We achieved a \ac{miou} score of \kittipercent~on the benchmark dataset, with a processing speed of \kittihz~\unit{\hertz}.
In addition, we evaluated our approach on our factory site dataset.
Our method achieved a \ac{miou} score of \fsdpercent~at \fsdhz~\unit{\hertz}.\\
\end{abstract}


\input{chapter/introduction}

\input{chapter/methods}

\input{chapter/results}

\input{chapter/conclusion}



\bibliographystyle{plain}
\bibliography{literature}


\end{document}

%% file: chapter/introduction.tex
\section{\large{Introduction}}
As \acp{agv} become increasingly prevalent in the industry, there is a growing need for methods to enable their safe operation.
Although small \acp{agv} such as flat load carriers and pallet lifters used indoors are already highly autonomous, counter-weight forklifts that operate primarily in mixed environments, \example{in- and outdoors}, are still human operated.
There are many challenging tasks to solve, such as free space and object detection, to operate the \acp{agv} safely.
Especially free space detection is crucial for subsequent tasks like path or task planning in warehouses.
Accurate free space detection allows for the planning of optimal trajectories and the prediction of operation time.
Additionally, it can provide additional information for object detection and tracking, which is necessary to detect obstacles and prevent potential collisions with other vehicles or pedestrians.
However, using \acp{agv} in warehouse environments introduces additional challenges that must be addressed.
These challenges include operating in highly dynamic indoor and outdoor environments and transferring between them.
The presence of unstructured areas, including load carriers and racks, without uniform lane markings, is an additional challenge.
Furthermore, the high angular velocities resulting from rear-wheel steering can lead to motion blur, further complicating free space detection.
When attempting to detect free space using \ac{lidar} sensors, additional challenges may arise from high reflecting surfaces, such as wet streets and low vertical resolution.

This paper presents a novel approach to addressing the challenges of free space detection during forklift navigation.
Our method uses a heuristic approach that leverages surface normals to create height change features to determine free space areas accurately.
Moreover, the presented method exhibits low latency, making it well-suited for real-time collision prevention in the demanding factory environment.

\subsection{State of The Art}
Free space detection is an essential task in robotics, particularly in the context of autonomous systems such as self-driving cars, drones, and logistics robots. This section presents a succinct overview of the state of the art in surface normal estimation, which our method requires, and research in free space detection that splits into heuristic and \ac{ml} based methods. A variety of techniques that have been presented in the literature are discussed, highlighting their strengths and limitations. However, the scope of this overview is limited to methods that utilize \ac{lidar} sensors, as this is the sensor employed in our presented method.
\subsubsection{Surface Normal Estimation}
Surface normal estimation in \ac{lidar} point clouds is mandatory for some approaches in downstream tasks like ground segmentation, scene understanding, collision avoidance and occlusion inference, scene reconstruction, localization, and many more. However, estimating surface normals from \ac{lidar} point clouds is a non-trivial task due to the usually unstructured nature of \ac{lidar} point clouds and sensor noise.
Approaches in prior work either rely on strong assumptions or come with a significant computational load. Rusu \cite{RusuDoctoralDissertation} shows that estimating a normal in a 3D point can be approximated by estimating a normal of a plane tangent to the surface that includes the point. In that case, the computational load equals solving a least-square plane fitting estimation to retrieve the tangent plane. Therefore, the solution for estimating the surface normal is reduced to analyzing the eigenvectors and eigenvalues of a covariance matrix created from the nearest neighbors of the query point. This involves finding the nearest neighbors, which is computationally heavy on unstructured point clouds. 
In \cite{2022.normal_transformer}, the authors use a combined approach between \ac{lidar} and RGB cameras to estimate surface normals through latent fusion in a neural network.
With LO-Net \cite{LO-Net}, the authors use a projection-aware representation of a \ac{lidar} point cloud to find neighboring points and compute the surface normal of a point as a weighted average between neighboring points. This approach requires computing a cross-product between the point of interest and each of its neighbors. The surface normals are further used for an odometry downstream task in their work.

\subsubsection{Heuristic Based Free Space Detection}
\label{sec:introduction_sota_heuristic}
In the ground detection field, heuristic approaches utilize rule-based filtering to distinguish ground points from non-ground points in point clouds. These methods are known for low processing times but can suffer from lower-quality ground detection. This paper compares our presented heuristic approach to some of these baseline methods.

The RANSAC~\cite{1981.ransac} algorithm is well-known and often used for finding planes in data. However, it is not optimal for extensive point cloud data as it repeatedly samples subsets to perform the plane fitting. Furthermore, this will only achieve good results on flat surfaces, as it cannot fit slopes or uneven terrain.
The authors of \cite{2010.linefit} propose a method in which they assume the point cloud's xy-plane forms a circle with an infinite radius. Afterward, they split the circle into segments, forming bins of points and sorting them by distance. Then, they fit lines that describe the ground in that particular bin, which allows them to assign each potential ground point to the closest line within a threshold. The method works well when points are close to the center of the point cloud, as the density of lines decreases with increasing distance. Anyway, it is a swift method.
In~\cite{2017.gpf}, a ground plane fitting method is introduced that performs a ground plane fitting to $N$ segments along the x-axis of the point cloud. To increase the processing speed of the plane fitting, they propose a height-based filtering that includes all points around the average lowest points of the point cloud within a certain threshold. With the segmented ground plane fitting, they can detect slopes and uneven terrain to a certain extinct.
In \cite{2021.r_gpf}, the authors propose a method to detect free space by encoding the point cloud and a prior map into volumes of interest. Afterward, they build region-wise bins in which they calculate occupancy descriptors. These descriptors are then used to fetch bins containing dynamic objects, which must be in contact with the ground. They then apply a \ac{rgpf} to detect the free space.
The Cascaded Ground Segmentation~\cite{2018.cascaded_seg} uses a two-stage filter process to detect free space. In the first step, an Inter-ring distance-based filter is applied to the laser scan, eliminating most non-ground points, followed by a multi-region plane fitting on the remaining potential ground points.
Lim~\etal also use plane fitting in their work \cite{2021.patchwork}, but extend it by splitting the point cloud into concentrically created bins. Combining the results allows them to perform multiple plane fittings for each bin and get a final free space estimation. It is also reasonably fast and can describe uneven terrain to a certain extent.

\subsubsection{ML Based Free Space Detection}
\label{sec:introduction_sota_ml}
Free space detection is commonly studied using machine learning algorithms, with image-based methods including semantic segmentation networks such as Mask RCNN~\cite{2017.mask_rcnn}, PointRend~\cite{2019.pointrend}, multitask CNNs~\cite{2019.multi-task_cnn}, and SNE~\cite{2020.sne}.
In addition, some methods perform free space detection on \ac{lidar} data, which utilize machine learning~\cite{2020.salsanet}. Furthermore, multi-sensor setup methods are widely used and show good results~\cite {2020.salsanet,2019.lidar-camera_fusion,2018.hybrid_crf}. Especially \cite{2022.2dpass}, which uses a fused training scheme, are among the top-scoring methods on the semantic KITTI benchmark.

Nagy~\etal propose a machine learning-based method for detecting free space in their paper~\cite{2021.nagy}, using \acp{cnn} trained on 2D panoramic images created from 3D point cloud data using the KITTI dataset~\cite{2013.kitti-dataset}.
The \acp{cnn} performs semantic segmentation on these panoramic images to classify each pixel into one of two classes: road and non-road.
The evaluation of the models shows that they fall below the top-scoring methods of the KITTI road benchmark~\cite{2013.kitti-road-benchmark}.
Their best model can create panoramic images with subsequent semantic segmentation at \SI[round-mode=places, round-precision=0]{17.24}{\hertz} processed on an NVIDIA\textsuperscript{\textregistered}~GPU.
The processing speed restricts the usage of the method. Furthermore, it makes it unattractive for highly dynamic environments, especially with sensors that incorporate \acp{lidar} with more than 64 layers, as this would decrease processing speed further.

Although most machine learning-based methods can achieve outstanding results on several benchmarks, these methods lack real-time capabilities, especially regarding large point clouds and processing without GPUs.

\subsection{Main Contributions}
In the state of the art, various approaches have been presented for surface normal estimation and free space detection, from heuristic or machine learning-based methods. 
The heuristics often cannot achieve such high quality as \ac{ml}-based methods but have higher run-times and no need for labeled ground truth data.
For \ac{ml}-based methods, on the other hand, labeled ground truth data is required for training, and the methods struggle to achieve real-time behavior. 
In this context, the main contributions of this paper can be summarized as follows:
\begin{itemize}
    \item A novel and fast method for estimating surface normals utilizing convolutional gradient filters, which is computationally very effective and results in real-time behavior of the free space detection.
    \item The presented algorithm is independent of labeled ground truth data, as it is not based on \ac{ml}, which saves labeling time and cost.
    \item Experimental evaluation of the presented method on two independent datasets with different \ac{lidar} sensors, environments, and installation positions and a benchmark against other heuristic methods for free space detection on one of them.
\end{itemize}

%% file: chapter/methods.tex
\section{\large{Methods}}
The following section contains an explanation of the methods used to solve the task of free space detection.
Our method is composed of three main steps. First, a spherical projection is described in \autoref{sec:spherical projection}, which is used to structure LiDAR point clouds. 
Each point in the point cloud is represented by its Cartesian coordinates $x$, $y$, and $z$ in sensor coordinates. 
The Ouster \ac{lidar}~\cite{ouster_os0} that is used, unlike others, creates an ordered array of points such that each neighboring point in the array corresponds to a neighboring point in the actual measurement.
Based on this, we calculate surface normals of the point cloud with a novel lightweight method as described in \autoref{sec:surface normal estimation}. Last, in \autoref{sec:height change}, a heuristic filter over the surface normals and the height of each point is used to identify points belonging to the ground and can be considered as free space.

\subsection{Spherical Projection}
\label{sec:spherical projection}
\begin{figure}[t] 
    \centering
    \includegraphics[width=\columnwidth]{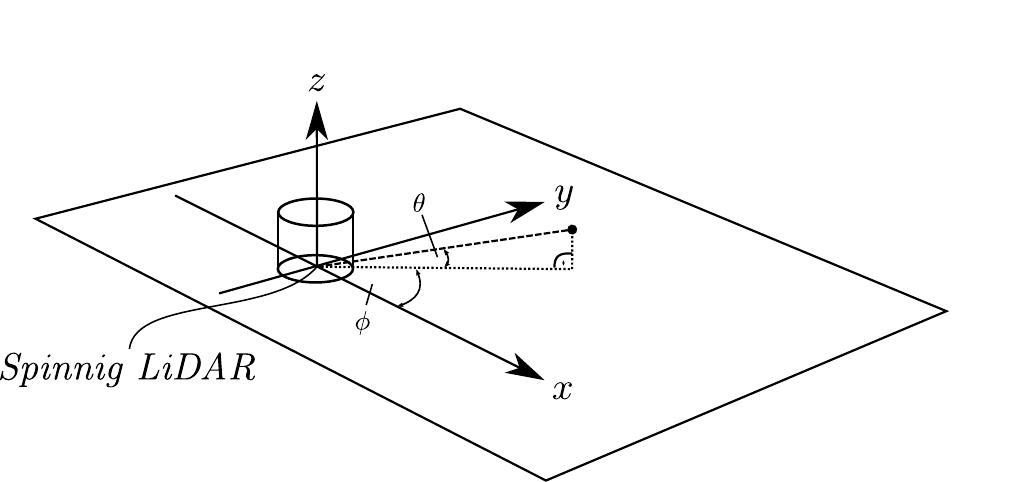}
    \caption{Coordinate system of a spinning LiDAR. $\phi$ as the azimuth angle and $\theta$ as an inclination angle from the $xy$-plane.}
    \label{fig:spinning_lidar}
\end{figure}
Our methods rely on a staggered image-like representation, in which neighboring pixels correspond to neighboring \ac{lidar} measurement rays and measured points, introduced by \mbox{Reichert}~\etal in \cite{2023.reichert}. We use a spherical image projection to obtain this representation. Specifically, the process involves converting Cartesian coordinates of the measurement points into spherical coordinates, as illustrated in \autoref{fig:spinning_lidar}. 
For each point in the point cloud, represented by its Cartesian coordinates $[x,y,z]^T$, we convert them into spherical coordinates represented by $[\phi, \theta, r]^T$. The azimuth $\phi$ corresponds to the angle of the point in the $xy$-plane, the inclination $\theta$ is the angle from the positive $z$-axis, and $r$ is the distance from the origin. This spherical projection captures the geometry of the sensor in a single image. We then use the following projection model to obtain $\vec{u}$ of a 3D point in a staggered spherical image representation:
\begin{equation}
\underbrace{\begin{bmatrix}
 u\\
 v\\
 1\\
\end{bmatrix} }_{\vec{u}}
=
\underbrace{\begin{bmatrix}
\frac{1}{\triangle \phi} &  0 &  c_{\phi}\\
 0 &  \frac{1}{\triangle \theta} &  c_{\theta}\\
 0 & 0 & 1\\
\end{bmatrix}}_{\mathbb{K}} \cdot \underbrace{\begin{bmatrix}
 \phi\\
 \theta\\
 1\\
\end{bmatrix}}_{\vec{x}}  
\end{equation}

\begin{figure}[t] 
    \centering
    \includegraphics[width=\columnwidth]{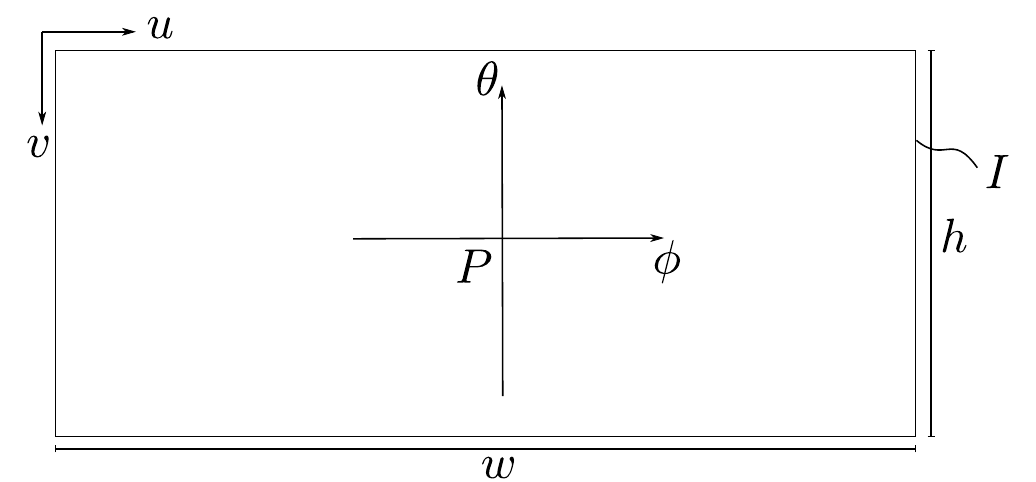}
    \caption{Spherical image $I$ with principal point $P$, height $h$, and width $w$.}
    \label{fig:spherical}
\end{figure}

Analogous to the projection model of pinhole cameras, the projection matrix $\mathbb{K}$ describes a discretization $\triangle \phi$, $\triangle \theta$ along the angles $\phi$, $\theta$ and a shift of the center coordinates $c_{\phi}$, $c_{\theta}$ defined by the height and width of the resulting image. Since the discretization can cause several points to be projected onto one pixel, we only use the points with the smallest Euclidean distance $r$ to the sensor. For a conventional spinning \ac{lidar} sensor, the image height $h$ and width $w$ will be equivalent to the number of layers and azimuth increments, respectively, as depicted in \autoref{fig:spherical}.
With the Ouster sensors used for this work, this operation is performed on-chip, which keeps the run-time low.
With the spherical projection of the ordered point cloud, it is possible to create a staggered image representation of the point cloud $I_{x,y,z}(u,v)=[x,y,z]^T$.

\subsection{Surface Normal Estimation}
\label{sec:surface normal estimation}
\begin{figure}[h]
    \centering
    \includegraphics[width=\columnwidth]{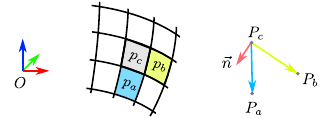}
    \caption{From left to right: Coordinate system of the sensor; The pixel $p_c$ is with its neighboring pixels $p_a$ and $p_b$ in the spherical projection; The corresponding 3D Cartesian points $P_c$, $P_a$, and $P_b$, along with the normal vector $\vec{n}$.} 
    \label{fig:normal}
\end{figure}

To calculate the normal vector of a surface in a 3D point, the typical approach is to use the cross product of two vectors that form a tangent plane in this 3D point. 
This paper presents a method for calculating surface normals based on the ordered point cloud.

Given the neighboring points $P_c$, $P_a$, and $P_b$ we can compute the surface normal at $P_c$ by forming the normalized cross product:
\begin{equation}
    \vec{n} = \frac{\overrightarrow{P_cP_b} \times \overrightarrow{P_cP_a}}{\norm{\overrightarrow{P_cP_b} \times \overrightarrow{P_cP_a}}_2}
    \label{eq:grad_u}
\end{equation}

We use the spherical image $I_{x,y,z}$ to find the neighborhood of a point in $I_{x,y,z}$. As shown in \autoref{fig:normal} the neighborhood of a pixel $p_c=(u,v)$, namely $p_b=(u+1,v)$ and $p_a=(u,v+1)$ correspond to neighboring points $P_c=I_{x,y,z}(u,v)$, $P_b=I_{x,y,z}(u+1,v)$, and $P_a=I_{x,y,z}(u,v+1)$ in the 3D point cloud. This allows us to use a directional derivative filter over the individual channels of $I_{x,y,z}$ to build the vector $\overrightarrow{P_cP_b}$ for every pixel position $(u,v)$:
\begin{equation}
\underbrace{\begin{bmatrix}
 I_x(u+1,v) - I_x(u,v)\\
 I_y(u+1,v) - I_y(u,v)\\
 I_z(u+1,v) - I_z(u,v)\\
\end{bmatrix}}_{
I_{\overrightarrow{P_cP_b}}(u,v)} = \begin{bmatrix}
(S_u * I_x)(u,v) \\
(S_u * I_y)(u,v) \\
(S_u * I_z)(u,v) \\
\end{bmatrix}
\label{eq:grad_img_u}
\end{equation}
With $S_u$ as a horizontal gradient filter and $*$ as the convolution operator. With $S_v$ as the vertical derivative filter we can build $\overrightarrow{P_cP_a}$:

\begin{equation}
\underbrace{\begin{bmatrix}
 I_x(u,v+1) - I_x(u,v)\\
 I_y(u,v+1) - I_y(u,v)\\
 I_z(u,v+1) - I_z(u,v)\\
\end{bmatrix}}_{
I_{\overrightarrow{P_cP_a}}(u,v)} 
= \begin{bmatrix}
(S_v * I_x)(u,v) \\
(S_v * I_y)(u,v) \\
(S_v * I_z)(u,v) \\
\end{bmatrix}
\label{eq:grad_img_v}
\end{equation}

\begin{figure}[t]
    \centering
    \includegraphics[width=\columnwidth]{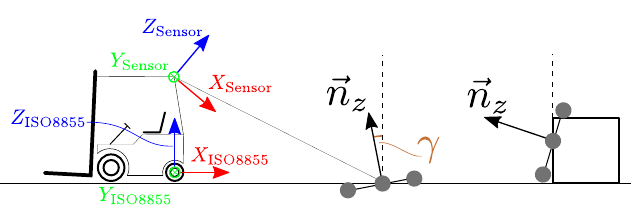}
    \caption{Each point (depicted in gray) is transformed from sensor coordinates to the forklift coordinate system. $\vec{n}_z$ shows the direction of the ground normal.}
    \label{fig:angles}
\end{figure}

 To obtain the surface normals $I_{\vec{n}}(u,v)$ we simply build the cross product over those vectors:
 
 \begin{equation}
    I_{\vec{n}}(u,v) = \frac{I_{\overrightarrow{P_cP_b}}(u,v) \times I_{\overrightarrow{P_cP_a}}(u,v)}{\norm{I_{\overrightarrow{P_cP_b}}(u,v) \times I_{\overrightarrow{P_cP_a}}(u,v)}_2}
    \label{eq:cross_img}
\end{equation}

\begin{figure*}[t] 
    \centering
    {%
      \includegraphics[width = \textwidth]{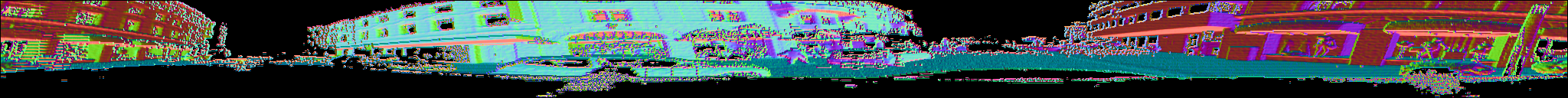}
      \includegraphics[width = \textwidth]{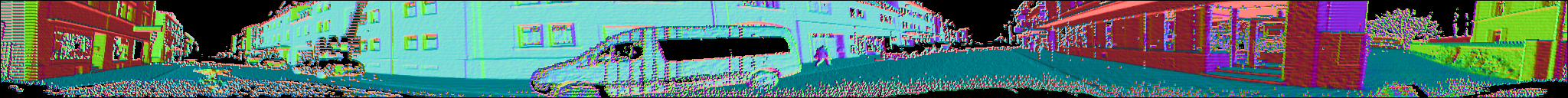}
      \includegraphics[width = \textwidth]{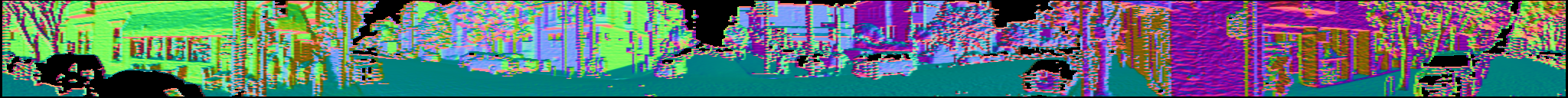}
      \includegraphics[width = \textwidth]{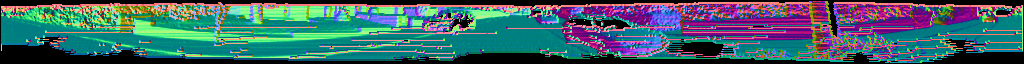}
    }
    \caption{Samples of our surface normal estimation for different sensors. From top to bottom: Ouster OS0 128x2048, Ouster OS1 128x2048, Ouster OS2 128x1024, Velodyne HDL 64x1024 (from KITTI). The surface normals' x, y, and z components are encoded in red, green, and blue. We can see the orientation and sign of the normals in the colors. Planar objects, such as the street or buildings, each have a uniform color.}
    \label{fig:surface_normal_samples}
\end{figure*}

To account for local noise in the point clouds and to achieve rotational symmetry, we use Scharr \cite{2000.scharr} filter to build the image gradients. 
In \autoref{fig:surface_normal_samples}, we show some samples of our surface normal estimation method for different LiDAR sensors. Our method utilizes six convolution operations, with two for each component along the x, y, and z axes. These convolution operations have a low computational load and are, therefore, very fast.

\subsection{Height Change Features}
\label{sec:height change}
The points in the point cloud and their corresponding pixels are transformed to the vehicle coordinate system defined in ISO8855 \cite{ISO8855} using a transformation matrix:

\begin{equation}
    I_{ISO8855|x,y,z}(u,v) = \begin{bmatrix}
        R|t
    \end{bmatrix}  I_{x,y,z}(u,v)
    \label{eq:transform}
\end{equation}

This transformation is crucial as it assumes that the immediate surroundings' ground is parallel to the vehicle's $xy$-plane coordinate system.
Furthermore, we assume that the immediate surrounding ground is flat, and the presented method is expected to identify any deviations from flatness in the remaining ground points.

\begin{figure}[t]
    \centering
    \includegraphics[width =\columnwidth]{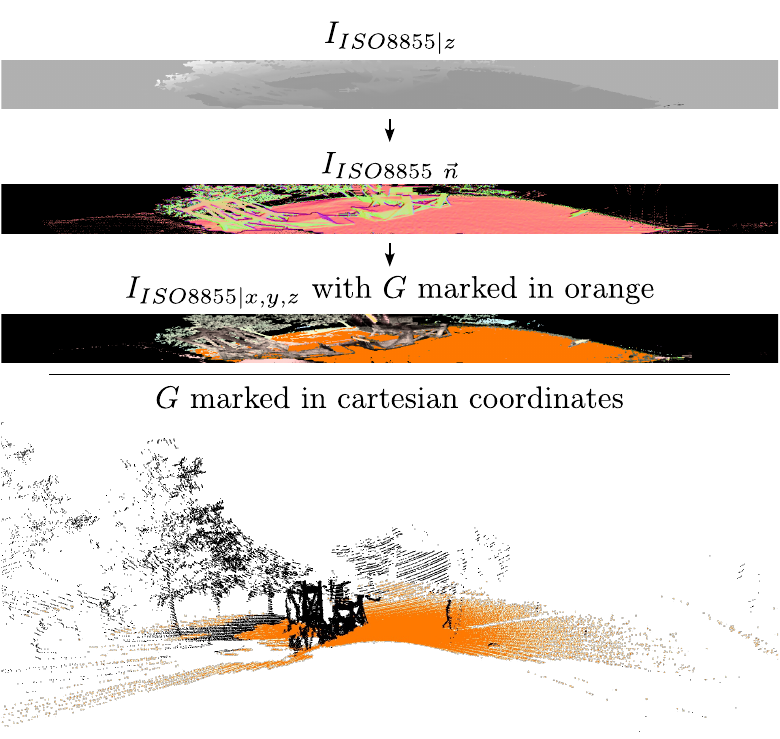}
    \caption{Pipeline of free space detection; First image shows $I_{ISO8855|z}$; Second image shows $I_{ISO8855~\vec{n}}$; Third image shows $I_{ISO8855|x,y,z}$ with the obtained set $G$ marked in orange; Fourth shows the back-projection from spherical image to cartesian coordinates}
    \label{fig:free_space_segmentation}
\end{figure}

To detect driveable areas (\thatis free space), we propose utilizing height change features: $cos(\gamma)(u,v)=I_{ISO8855~\vec{n}|z}(u,v)$ ($\gamma$: see \autoref{fig:angles}). $\gamma$ is the angle between the normal $I_n(u,v)$ and the z-axis in the ISO8855 vehicle coordinate system. For drivable areas, it should be close to 0 degrees meaning that $cos(\gamma)$ is close to 1.
The height change features are then utilized to separate the points $C = \left\{(u,v): u \in [0,\ w[,\ v \in [0,\ h[ \right\}$ into a set $V$ (verticals) and $\overline{V}$ (not verticals), representing the remaining points:

\begin{equation}
    V = \left\{(u,v) \in C \left\vert 0.90 \leq cos(\gamma)(u,v) \leq 1.0 \right.\right\}
    \label{eq:possible_ground_points}
\end{equation}
The lower boundary of \autoref{eq:possible_ground_points} is an empirically obtained hyper-parameter.
The resulting set $V$ contains all points where $\vec{n}$ points upward. 

Not all points of $V$ belong to the ground plane (some points may lie on a plane parallel to the ground plane), so we need further filtering. However, we assume that points on the ground plane are a majority of $V$. Therefore, we apply a statistical filter with a $\filtersigma\sigma$ distribution to $V$ with \eqref{eq:ground} to get the ground points $G$, where $\sigma$ is the standard deviation and $\mu$ the mean of $I_{ISO8855|z}(u,v)$ of $V$.

\begin{equation}
    G =
    \left\{ (u,v) \in V \left\vert 
        \begin{array}{l}
            \lvert I_{ISO8855|z}(u,v) - \mu \rvert < \sigma
        \end{array}
    \right.
    \right\}
    \label{eq:ground}
\end{equation}

$G$ contains all ground points not occupied with objects, while $OG = \overline{G}$ contains all off-ground points.
\autoref{fig:free_space_segmentation} shows the whole pipeline from the $I_z$ to the ground points $G$ (depicted in orange) of one \ac{lidar}.
Finally, $G$ is interpreted as the free space on which the forklift can move freely and without colliding with any object.

%% file: chapter/results.tex
\section{\large{Results}}
In this section, we present the results of our experimental evaluation of the presented method for free space detection that considers height change features. Our method was designed to accurately and efficiently segment free space in environments with complex structures and terrain variations. To validate the effectiveness of our approach, we evaluated it on two datasets: the Semantic KITTI dataset~\cite{2019.semantic_kitti}, which is an extension of the original KITTI Vision Benchmark~\cite{2012.kitti} and a Factory Site dataset that we obtained on a real-world factory site containing indoor and outdoor scenes. 
We employed the \ac{miou} metric to assess the segmentation accuracy of our method.
This metric measures the overlap between the predicted ground pixels and the pre-labeled ground truth pixels used for evaluation.
A higher value of \ac{miou} indicates a closer alignment between the predicted and actual ground, which is a better detection.

\subsection{Semantic KITTI evaluation}
Our presented method was evaluated using the annotated LiDAR data provided by the Semantic KITTI dataset.
As the dataset lacks a dedicated label for free space, we aggregated the labels for \textit{road}, \textit{parking}, \textit{sidewalk}, \textit{other-ground}, and \textit{lane-marking} and treated all pixels belonging to these classes as free space.
Additionally, to ensure consistency in our evaluations, we used the same set of parameters optimized for the Factory Site dataset without any modifications tailored to the Semantic KITTI dataset.

We benchmark our method against six heuristic baseline methods also applied to the semantic KITTI dataset for comparability. The benchmark results are shown in \autoref{tab:results_kitti}. The speeds of the baseline methods were measured in \cite{2022.patchwork++_benchmarks} on an Intel\textsuperscript{\textregistered}~Core\texttrademark~I7 CPU. To ensure comparability, we processed our method using the same CPU.

\begin{table}[ht]
    \centering
    \captionsetup{justification=centering}
    \caption{Semantic KITTI Results.\\ Hardware: Intel\textsuperscript{\textregistered}~Core\texttrademark~i7 CPU}
    \label{tab:results_kitti}
    \begin{tabular}{c|c|c}
         Algorithm                                                      & mIoU [\%] $\uparrow$  & Speed [Hz]  $\uparrow$  \\
         \hline
         R-GPF~\cite{2021.r_gpf}                                        & 32.30                 & \num[round-mode=places, round-precision=0]{35.30} \\
         RANSAC~\cite{1981.ransac}                                      & 37.60                 & \num[round-mode=places, round-precision=0]{15.43} \\
         CascadedSeg~\cite{2018.cascaded_seg}                           & 37.87                 & \num[round-mode=places, round-precision=0]{13.07} \\
         GPF~\cite{2017.gpf}                                            & 41.12                 & \num[round-mode=places, round-precision=0]{29.72} \\
         Patchwork~\cite{2021.patchwork}                                & 41.32                 & \num[round-mode=places, round-precision=0]{43.97} \\
         LineFit~\cite{2010.linefit}                                    & 43.05                 & \num[round-mode=places, round-precision=0]{58.96} \\
         \hline
         \textbf{H-FSD (ours)}                                          & \textbf{\kittimiou}   & \textbf{105}
    \end{tabular}
\end{table}
The current state-of-the-art \ac{ml} based model for single input semantic segmentation on the Semantic KITTI dataset is 2DPASS~\cite{2022.2dpass}, which achieves a \ac{miou} of \SI{85.7}{\percent} among the classes we consider, with a processing speed of \SI{16}{\hertz}.
Our presented approach can reach a \ac{miou} of \kittipercent~at a significantly faster processing time of \kittihz~\unit{\hertz}, which makes it suitable for real-time applications.

\subsection{Factory site dataset evaluation}
The experiments were conducted at a factory site, utilizing an outdoor dedicated test site and an indoor test area within the production hall.
The indoor scenes include mostly flat grounds with different color markings and partly lane markings, while the outdoor scenes include longitudinal slopes up to \SI{12}{\percent}, mostly on asphalt roads.
We utilized two Ouster OS0 \ac{lidar} sensors with a vertical resolution of \num{128} layers and a horizontal resolution of \num{2048}, providing a wide vertical \ac{fov} of \ang{90} with $\pm \ang{45}$ coverage from the horizon and \ang{360} horizontal \ac{fov}.
\autoref{fig:free_space_segmentation} shows a sample of the outdoor test site processed with the height change features of one of the Ouster \ac{lidar}.
The results are summarized in \autoref{tab:fsd_results}.
The \ac{miou} achieved on the Factory Site dataset is higher, which can be attributed to the increased vertical resolution resulting from the doubled layer count compared to the Velodyne \ac{lidar}. The Factory Site dataset also contains more artificial structures like roads, buildings, and racks which produce noiseless surface normals. In contrast, the KITTI dataset includes a lot of vegetation, leading to noisy surface normals. Therefore, more outliers appear, resulting in a lower \ac{miou}.
However, this improvement comes at the cost of reduced processing speed, with the OS0 \ac{lidar} operating at nearly half the speed of the Velodyne.
We can reach \fsdhz \unit{\hertz}, which is enough to enable the real-time application.

\begin{table}[ht]
    \centering
    \caption{Factory Site Dataset Results}
    \label{tab:fsd_results}
    \begin{tabular}{c|c|c}
         Dataset        & Semantic KITTI        & Factory Site              \\
         \hline
         \# Pointclouds & \kittisamples         & \fsdsamples               \\
         \ac{lidar}     & 1x Velodyne HDL-64E   & 2x Ouster OS0 128    \\
         \ac{miou} & \kittipercent         & \fsdpercent                  \\
         Avg. Speed   & \kittihz \unit{\hertz}         & \fsdhz \unit{\hertz}~per \ac{lidar}
    \end{tabular}
\end{table}



One advantage of our presented method is that it does not require training data or machine learning models. This eliminates the need for time-consuming training steps and reduces the reliance on ground truth data, which can be costly and difficult to obtain in some scenarios. Instead, our method relies on simple heuristic rules based on height change features, making it easy to implement and interpret.

Overall, our results demonstrate the effectiveness of our presented heuristic method for free space detection, which achieves accurate and efficient segmentation of ground points without the need for training data or machine learning models.

%% file: chapter/conclusion.tex
\section{\large{Conclusion and Future Work}}
This paper introduced a novel heuristic method for efficient free space detection by leveraging a surface normal estimation technique that incorporates a spherical projection of \ac{lidar} point clouds. The presented method is evaluated on the widely-used Semantic KITTI dataset and is compared against several heuristic baseline methods. Notably, the method demonstrates low run-time, which can potentially be further accelerated by implementing GPU processing instead of CPU. Future research directions include slope handling and exploring object detection techniques that utilize the precise position information of obstacles obtained from the $OG$ set. We also assume that our fast and easy surface normal estimation method can be useful in a collection of tasks that depend on surface normals like point-to-plane-based odometry.